\pdfoutput=1

\documentclass[11pt]{article}

\usepackage{authblk}

\usepackage[preprint]{acl}

\usepackage{times}
\usepackage{latexsym}

\usepackage[T1]{fontenc}

\usepackage[utf8]{inputenc}

\usepackage{microtype}

\usepackage{inconsolata}

\usepackage{tipa}
\usepackage{graphicx}
\graphicspath{ {./figures/} }

\usepackage{booktabs}

\usepackage{tikz}
\usetikzlibrary{shapes, arrows.meta, positioning}
\usepackage{amsmath}

\title{Generating Feature Vectors from Phonetic Transcriptions in Cross-Linguistic Data Formats}

\author[1]{\textbf{Arne Rubehn}}
\author[1]{\textbf{Jessica Nieder}}
\author[2]{\textbf{Robert Forkel}}
\author[1]{\textbf{Johann-Mattis List}}
\affil[1]{Chair for Multilingual Computational Linguistics, University of Passau \protect\\ Passau, Germany \protect\vspace{3mm}}
\affil[2]{DLCE, MPI-EVA \protect\\ Leipzig, Germany}

\begin{document}
\maketitle
\begin{abstract}

When comparing speech sounds across languages, scholars often make use of feature representations of individual sounds in order to determine fine-grained sound similarities. Although binary feature systems for large numbers of speech sounds have been proposed, large-scale computational applications often face the challenges that the proposed feature systems -- even if they list features for several thousand sounds -- only cover a smaller part of the numerous speech sounds reflected in actual cross-linguistic data. In order to address the problem of missing data for attested speech sounds, we propose a new approach that can create binary feature vectors dynamically for all sounds that can be represented in the the standardized version of the International Phonetic Alphabet proposed by the Cross-Linguistic Transcription Systems (CLTS) reference catalog. Since CLTS is actively used in large data collections, covering more than 2,000 distinct language varieties, our procedure for the generation of binary feature vectors provides immediate access to a very large collection of multilingual wordlists. Testing our feature system in different ways on different datasets proves that the system is not only useful to provide a straightforward means to compare the similarity of speech sounds, but also illustrates its potential to be used in future cross-linguistic machine learning applications.

\end{abstract}

\section{Introduction}
The past two decades have seen a drastic increase in standardized datasets in historical linguistics and linguistic typology which are available in both human- and machine-readable form \citep{dellert2020northeuralex,Skirgard2023,ASJP-16.0.0}. 
With Lexibank (\url{https://lexibank.clld.org}, \citealt{list2022lexibank}), a large collection of comparative wordlists has been compiled in which word forms from various independently published datasets are standardized along three dimensions, including (1)~the languages in which they occur, (2)~the concepts which they express, and (3)~the sounds that constitute them. 
Lexibank is a result of the more general Cross-Linguistic Data Formats initiative (CLDF, \url{https://cldf.clld.org}, \citealt{forkel2018cross}), which aims to unify several kinds of cross-linguistic data (wordlists, typological datasets, interlinearglossed texts) by proposing an exchange format along with guidelines and recommendations for standardization.
 
Sounds in Lexibank are represented in a unified transcription system, proposed as part of the Cross-Linguistic Transcription Systems reference catalogue (CLTS, \url{https://clts.clld.org}, \citealt{CLTS}) that can be considered a large standardized subset of the International Phonetic Alphabet (IPA, \citealt{IPA1999}). The CLTS system tries to handle as much of the variation observed in phonetic transcriptions as possible, using a dynamic method that parses phonetic transcriptions in a given transcription system and derives features from individual symbol combinations
\citep{anderson2018cross}. 
The feature system underlying these transcriptions has been designed in a pragmatic way that would allow to capture as much of the graphical variation in using the IPA (and other transcription systems) as possible \citep{Anderson2023}. As a result, the system is powerful in parsing phonetic transcriptions -- specifically those represented in IPA -- but it is not particularly useful to compare speech sounds with respect to their \emph{similarity} (be it acoustic or articulatory or a combination of both). 

Thus, while the CLTS system does its job in helping to standardize phonetic transcriptions in an unprecedented way, as witnessed by the Lexibank collection (and numerous additional CLDF wordlists that have been published in the past years), it falls short in providing a reliable means to compare individual sounds for their similarity.

In this study, we present a very straightforward approach to convert the CLTS feature system to a vector representation. This approach takes CLTS feature bundles as input and converts them into binary feature vectors that can be used for various downstream tasks in computational phonology, computational historical linguistics, and computational linguistic typology.

\section{Background}

Modeling speech sounds as bundles of distinctive features can be seen as the most typical and straightforward way in phonology and comparative linguistics to compare the similarity of speech sounds. It is therefore not surprising, that phonological features play a crucial role in a number of different approaches, ranging from historical language comparison \citep{kondrak2000new} over dialectology \citep{nerbonne1997measuring,hoppenbrouwers2001} and phonological rule induction \citep{gildea1996learning} to child language acquisition \citep{somers1998similarity}. 
Representing sounds with the help of  features can also enhance the performance of automatic speech recognition \citep{metze2007using} and transliteration \citep{tao2006unsupervised,yoon2007multilingual}, as well as automatic phonetic transcriptions from text and named entity recognition \citep{mortensen2016panphon}. Related studies have additionally demonstrated that meaningful phone(me) embeddings can be learned from distributional properties \citep{silfverberg2018sound,sofroniev2018phonetic}.

Beyond these mostly implicit uses of phonological features, there exist several frameworks with the explicit purpose of modeling the interactions between sounds and their respective features. Computational tools for analyzing phoneme inventories in terms of phonological features and natural classes are available in the form of web applications \citep{steel2017featurize} or downloadable programs \citep{vanvugt2021pheatures}. However, these tools by default cover rather small inventories of fairly common sounds and are often even only designed for individual languages.

There is a small number of datasets that map a large number of sounds to a feature representation, aiming to cover a substantially large amount of speech sounds in order to be applicable in cross-linguistic studies. PanPhon \citep{mortensen2016panphon} defines feature representations for approximately 5,000 sounds, the similar but smaller framework DistFeat \citep{tresoldi2020model} spans roughly 500 sounds. 
Phonotacticon \citep{joo2023phonotacticon}, a typological resource for phonotactics, extends PanPhon to around 20,000 distinct speech sounds. PHOIBLE \citep{moran2019phoible}, a database covering various phoneme inventories, is equipped with feature definitions as well, covering all  3,000 distinct sounds attested in the data in its latest version. Finally, the Python package \texttt{ipasymbols} \citep{hamster2022everybody} is designed to query IPA symbols by their articulatory properties, but is not equipped with phonological features and is currently (v.0.1.0) limited to only 179 sounds.

While all feature collections are much larger than the earlier feature collections that phonologists proposed for individual languages, reflecting the trend towards cross-linguistic approaches that allow for a comparison across multiple languages, all feature collections are \emph{fixed sets of sounds}, lacking a dynamic component. This limits their potential when applying them to newly compiled datasets, since whenever a 
sound in a given dataset is not attested in the feature systems, users would have to add it or to label it as missing data.

While this may seem to reflect a minor problem, it has grown into a major obstacle for many concrete applications in computational comparative linguistics, since practical experience in working with concrete language data clearly shows that meeting unobserved sounds when turning to new datasets is rather the rule than the exception (see the observation in \citealt{Moran2012}, that the overall number of distinct speech sounds seems to increase almost constantly, albeit slowly, when new data are added to the sample). 
One way to avoid the problem of observing missing sounds is to arbitrarily extend mappings from IPA transcriptions of speech sounds to feature mappings in a systematic way, as exemplified by 
the extended system proposed by the Phonotacticon, with 20,000 distinct sounds, of which only a couple of hundred sounds occur in the final database. 

An alternative, more robust approach, specifically important for data standardized in CLDF, would take the pragmatically oriented non-binary features provided by the CLTS system as a starting point and convert them to a binary vector representation.



\section{Materials and Methods}

\subsection{Materials}

The starting point of our approach is the CLTS reference catalogue, which links feature descriptions of speech sounds in the style of the IPA to different transcription systems and datasets. While the CLTS website presents a list of about 8,000 distinct speech sounds that are linked to various datasets, including PHOIBLE and PanPhon, the system that generates the website is dynamic, with only a couple of hundred base sounds being defined explicitly. The rest of the sounds is generated from sound transcriptions mainly provided in the International Phonetic Alphabet. The dynamic system underlying the CLTS reference catalogue can be accessed with the help of a Python API (\url{https://pypi.org/projects/pyclts}, \citealt{PyCLTS}, see \citealt{anderson2018cross} for the details regarding the algorithm used by the API). 
As a result, IPA strings that are not directly represented in the system can be processed, as long as they conform to IPA standards (broadly defined by CLTS). The CLTS system parses sounds in two ways, taking a phonetic transcription (typically provided in IPA) as starting point, or starting from the typical name of a speech sound, as they are also defined by the IPA. For example, \textipa{[p]} would be described as the `voiceless bilabial stop consonant', yielding the descriptive feature set (`voiceless', `bilabial', `stop', `consonant'). 
For the conversion of sound transcriptions accepted by CLTS to binary feature vectors, we use the feature bundle representation rather than the phonetic transcription as our starting point.

In order to evaluate the usefulness of the binary feature vectors derived from CLTS, we use the Lexibank database, since it provides a large collection of wordlists that conform to the standard defined by CLTS.

As of version 1.0 \citep{Lexibank} Lexibank is available in an aggregated form in which all datasets that are sufficiently standardized  -- with all sounds being interpretable by the dynamic CLTS system -- are assembled in a single CLDF dataset that can be parsed and processed in various ways, including SQLite (see \citealt{List2023taste}) or Python (using the CL Toolkit package, see \citealt{CLToolkit}, \url{https://pypi.org/project/cltoolkit}). In this form, Lexibank covers wordlists of at least 80 distinct words for about 2,000 distinct language varieties.

\subsection{Methods}
\subsubsection{Feature System}\label{sec:feature-definitions}

We define a classical feature space of 39 binary phonological features that can be present (1) or absent (-1), or non-applicable (0). Strictly speaking, we therefore employ a notion of ternary, rather than binary features, since there are three instead of two possible values. However, this is merely an explicit formalization of the way that binary features are usually treated in phonology: Not all features can apply to all kinds of sounds. It is therefore necessary to distinguish absent from non-applicable features by assigning them different numerical values. To illustrate this, consider the feature \texttt{[±strident]} which only applies to fricatives and affricates \citep{zsiga2013sounds} -- it is worthwhile to distinguish non-strident fricatives (which \textit{could} be strident) from other sounds that do not have this feature at all. This notion of applicability is frequently found in the literature, and formally makes these systems ternary rather than binary. Keeping this in mind, we will still refer to this feature system as binary, given that it is the commonly used term to describe this kind of feature systems.

The majority of the features we define constitutes a fairly well established standard inventory, where we strictly follow the definitions by \citet{zsiga2013sounds}. Nevertheless, to be able to cover a comprehensive range of sounds, some additions to the feature inventory were required. We incorporate three additional features \texttt{[velaric], [hitone]}, and \texttt{[hireg]} from \citet{mortensen2016panphon} to handle clicks and tones. Clicks are assigned \texttt{[+velaric]} on top of their other features that are derived from their analogous pulmonic stops. The tonal features \texttt{[hireg]} and \texttt{[hitone]} refer to the broader register, and the more narrow tone quality within the register -- both the high tone [\textsuperscript{5}] and the mid-high tone [\textsuperscript{4}] belong to the high register and are therefore \texttt{[+hireg]}, and within that register, [\textsuperscript{5}] is the higher tone and is therefore \texttt{[+hitone]} (whereas [\textsuperscript{4}] is \texttt{[-hitone]} analogously). Since we do not want to assign the feature value of 0 (non-applicable) to tonal features in tones, however, these two features only yield 4 possible combinations, insufficient for encoding the canonical 5 tones. We therefore introduce the supplementary feature \texttt{[loreg]} as a logical counterpart to \texttt{[hireg]}, covering the low and the mid-low tones.

Furthermore, we employ three additional features to represent complex tones. Tonal features in CLTS are based on a rather schematic representation of Chao's numeral coding of complex tones in Chinese dialects \citep{Chao1930}. Thus, the tone [\textsuperscript{214}] is labeled as \texttt{"contour from-mid-low via-low to-mid-high tone"} in the CLTS name space, with the feature value \texttt{"from-mid-low"} (from the feature \texttt{[start]}) -- representing the number {\textsuperscript{2}}, \texttt{"via-low"} (from the feature \texttt{[middle]}) representing {\textsuperscript{1}}, and \texttt{"to-mid-high"} (from the feature \texttt{[end]}) representing {\textsuperscript{4}} directly, while the feature value \texttt{"contour"} (from the feature \texttt{[contour]} adds additional information that tells us that we are dealing with a contour tone. In the same way, CLTS assigns the values \texttt{"rising"} and \texttt{"falling"} to tones like [\textsuperscript{15}] and [\textsuperscript{51}] respectively. In the vector representation, the contour of a tone is directly translated to the features \texttt{[contour, rising, falling]} based on their respective CLTS feature. Contour tones with both rising and falling parts additionally receive \texttt{[+rising]} or \texttt{[+falling]}, indicating the interval between the initial and the middle segment. All complex tones inherit the features \texttt{[hireg, hitone, loreg]} from their initial segment. The example tone [\textsuperscript{214}] is therefore represented as \texttt{[-hireg, +hitone, +loreg, +contour, -rising, +falling]}.

Finally, seven more features are defined for representing diphthongs, namely \texttt{[backshift, frontshift, opening, closing, centering, longdistance, secondrounded]}. These features are used to model the diphthong's trajectory across the vowel space \citep[for a more detailed description see][41-43]{rubehn2022feature}. Additionally, each diphthong is assigned the simple vowel features of its initial segment.


Due to its flexibility, the present system is highly customizable and can be used to generate different feature systems as well. Users can easily define their own feature inventories and mappings, according to their individual needs. The workflow presented in the following section is not dependent on the specific feature system and definitions that we suggest here.

\subsubsection{Workflow}

\begin{figure}[tbh]
\centering
\begin{tikzpicture}[auto, node distance=1.5cm]
\tikzstyle{block} = [rectangle, draw, fill=blue!20, text width=10em, text centered, rounded corners, minimum height=4em]

\node [block, label={[align=left,anchor=east]left:[p]}] (string) {String Representation};
\node [block, below=of string, label={[align=left,anchor=east]left:voiceless \\ bilabial \\ stop}, fill=green!20] (features) {Descriptive Features};
\node [block, below=of features, label={[align=left,anchor=east]left:$
  \left[\begin{matrix}
    -\text{son} \\
    -\text{cont} \\
    +\text{lab}
  \end{matrix}\right]$}, fill=red!20] (vector) {Vector};

\draw [->] (string) -- node {\small{CLTS}} (features);
\draw [->] (features) -- node {\small{Hierarchical Vector Mapping}} (vector);

\end{tikzpicture}
\caption{Workflow of vector creation.}
\label{fig:workflow}
\end{figure}
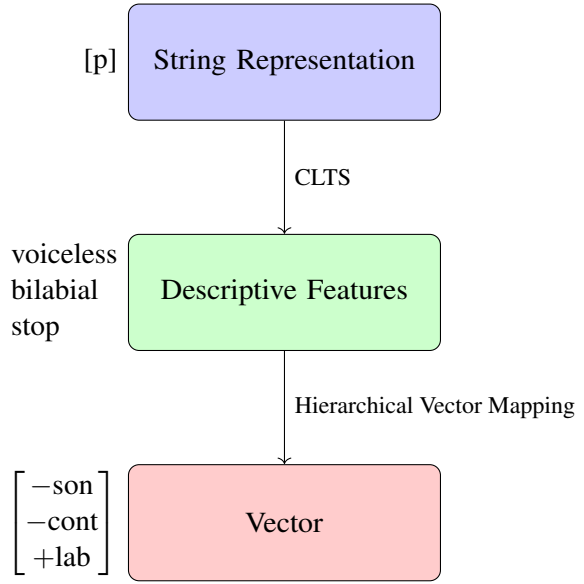

Our system generates binary feature vectors for any sound based on its feature set assigned by CLTS. Again, consider the example \textipa{[p]}: The method does not depart from the string \texttt{"p"}, but from the feature set (`voiceless', `bilabial', `stop', `consonant') that can easily be obtained from CLTS. The general workflow for generating binary feature vectors is outlined in Figure \ref{fig:workflow}.

Underlying our system is a simple dictionary structure which maps triplets of CLTS feature values, CLTS feature domains, and binary feature representations onto each other. The feature value `stop' for example would be linked to the domain `manner' (of articulation in consonants) and to the binary feature representation \texttt{[-son,-cont]}.

As a starting point, a zero vector (with the value 0 at every position) of the size of the defined feature inventory is instantiated, with every position of the vector corresponding to exactly one binary feature. This vector then is modified by subsequently processing the features in the CLTS feature set, with the corresponding binary features overwriting the current value in the vector.

A core principle for the successful modification of feature vectors is that we sort the CLTS features by a \textit{hierarchy} of concreteness that determines the order in which those features are processed. This hierarchy states that the least specific features get processed first, and the most specific ones get processed last. In the concrete example of \textipa{[p]}, that means that `consonant', being the least specific feature, is processed first.

This notion of hierarchy is necessary for handling conflicting feature mappings, since we deliberately allow for values in the vector to be overwritten by features that are processed later. To exemplify this, consider
the `devoiced voiced labio-dental fricative' \textipa{[\r*v]}: The descriptor `voiced' maps to \texttt{[+voice]}, whereas `devoiced' naturally corresponds to \texttt{[-voice]}. However, since diacritics modify the base sound, they should take precedence over it, and the correct feature that should be assigned is \texttt{[-voice]}. This is ensured by processing the features according to the hierarchy, which states that the modification `devoiced' is more concrete and should therefore be applied \textit{after} the regular phonation feature `voiced', and can thus overwrite the previously assigned \texttt{[+voice]} with \texttt{[-voice]}.

This notion of hierarchy also conveniently allows for the usage of \textit{default values} that can define which features apply at which representation level. This is important since we distinguish between non-applicable (0) and absent (-1) features, as discussed in Section \ref{sec:feature-definitions}. We can therefore define which set of binary features always applies to a certain group of sounds by assigning a default value to ensure that applicable features have a non-zero value. For example, the feature \texttt{[±lab(ial)]} must be defined for all consonants, which is assured by mapping the CLTS feature `consonant' to \texttt{[-lab]}. If the consonant is actually labial, this feature will be overwritten with \texttt{[+lab]}, since the place of articulation is always applied after the sound type. So instead of exhaustively defining \texttt{[-lab]} for every non-labial place of articulation, we can just define it as a default value for the CLTS feature `consonant' instead. This corresponds to the reading that every consonant is \texttt{[-lab]} (by default), unless specified otherwise.

The majority of sounds can be handled by this straightforward workflow of hierarchically mapping CLTS features to their binary feature representations. However, there are a few more complex cases that require an extra processing step. For example, the glottal stop \textipa{[P]} has the binary feature \texttt{[+cg]} (`constricted glottis') -- however, this feature neither corresponds to `glottal', nor to `stop'. It is therefore the combination of both `glottal' and `stop' that triggers \texttt{[+cg]}. The system therefore uses a second dictionary that allows for the definition of joint feature mappings, where a binary feature definition is conditioned by a certain \textit{combination} of CLTS features.

Complex sounds that can alternatively be analyzed as two segments -- diphthongs and consonant clusters -- pose a similar challenge. For these cases, CLTS provides the means of analyzing its individual constituents: The consonant cluster \textipa{[kp]} can be split into \textipa{[k]} and \textipa{[p]}. The system uses this function to generate separate feature vectors for the two individual sounds, which then are combined by assigning the union of positive features to the joint vector. The resulting feature vector for \textipa{[kp]} therefore contains all positive features that are attributed to either \textipa{[k]} or \textipa{[p]}. 

In a similar fashion, feature vectors for diphthongs are based on their initial segments. For example,  [\textipa{aI}] inherits its monophthong vowel features from the feature definitions for \textipa{[a]}. The additional diphthong features, that indicate the trajectory of the diphthong, are assigned based on joint feature definitions: The combination of the CLTS features (`from\_open', `to\_near-close') maps (among others) to the binary feature \texttt{[+closing]}.

\subsection{Implementation}
The approach is implemented in the form of a Python package (\texttt{soundvectors}) that takes as input the canonical names consisting of feature values that CLTS generates dynamically for speech sounds in standard IPA transcription and can be applied in combination with CLTS and the \texttt{pyclts} package, as well as with the \texttt{linse} package that offers non-generative access to a larger selection of speech sounds covered by CLTS (\url{https://pypi.org/project/linse}, \citealt{List2024TBLOGb}), but also independently of these packages, as long as the feature names follow the CLTS standards. The package along with the data on which it was tested is available from the supplementary material accompanying this study.

\section{Evaluation}
We test the usefulness of our proposed system by (1)~investigating the vector similarities for common sounds by calculating cosine similarities and visualizing them with heatmaps, (2)~employing techniques for dimensionality reduction to visualize the relationships between sounds, (3)~mapping the CLTS sound inventory to binary vectors and analyzing the resulting equivalence classes, and (4)~investigating the power of the system to distinguish speech sounds observed in phonetically transcribed wordlists.

\subsection{Vector Similarities}

\begin{figure}[tb]
    \centering
    \includegraphics[width=0.5\textwidth]{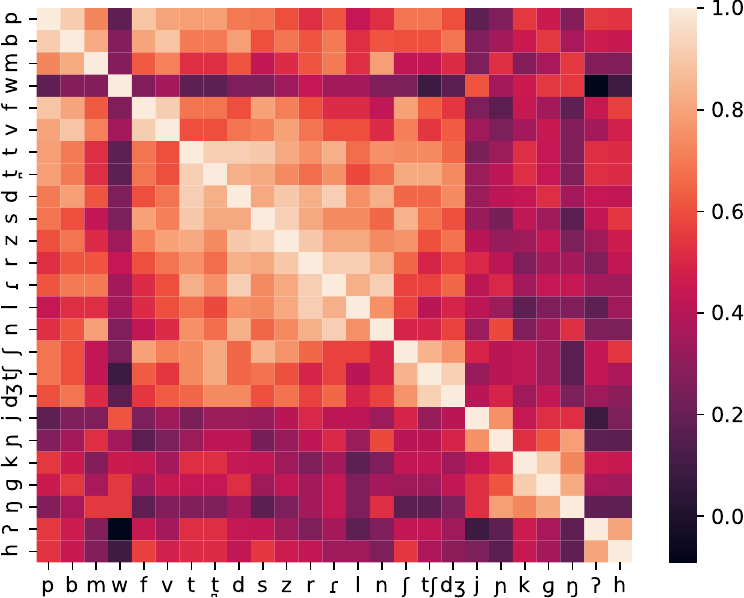}
    \caption{Cosine similarities between consonant vectors generated with our model.}
    \label{fig:heatmap-cons}
\end{figure}

\begin{figure}[tb]
    \centering
    \includegraphics[width=0.5\textwidth]{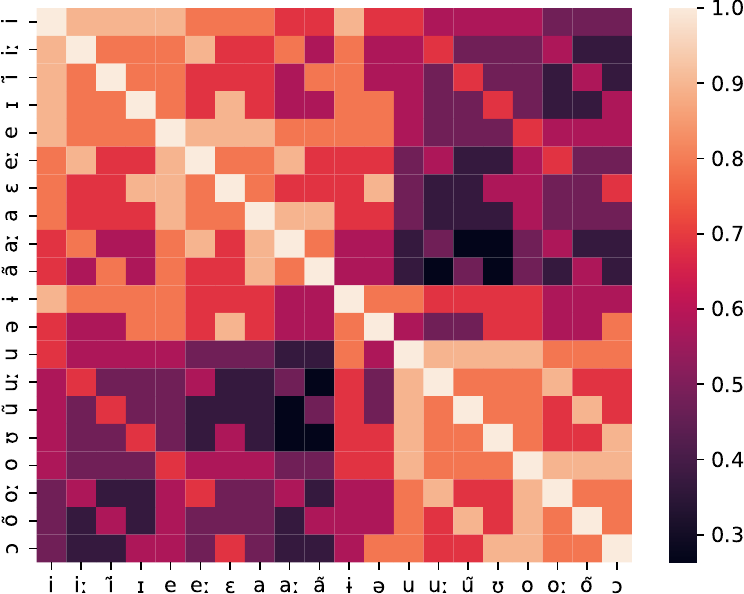}
    \caption{Cosine similarities between vowel vectors generated with our model.}
    \label{fig:heatmap-vow}
\end{figure}


To test how well our system analyzes a representative sample of common sounds, we take 25 most common consonants and the 20 most common vowels from Phoible 2.0 \citep{moran2019phoible} and use heatmaps to visualize the cosine similarities similarities of their respective feature vectors (Figures \ref{fig:heatmap-cons} and \ref{fig:heatmap-vow}).
The heatmaps were generated with the Python library Seaborn \citep{waskom2021seaborn}, with lighter colours representing higher similarities and darker colours representing lower similarity scores. The sounds are ordered by their primary place of articulation, starting from the front and moving to the back of the mouth. 
As can be observed in Figure \ref{fig:heatmap-cons}, the manner of articulation and the phonation have a clear impact on the similarity of consonantal feature vectors: \textipa{[p]} is therefore much more similar to \textipa{[k]} than to \textipa{[N]}. The glides \textipa{[w]} and \textipa{[j]} are strikingly dissimilar to the rest of the consonants, showing their well-known intermediate role in between consonants and vowels.

Figure \ref{fig:heatmap-vow} shows that the vowel space follows a strong division into two clusters which correspond to front and back vowels, with \textipa{[a]} being considered a front vowel according to the IPA nomenclature. This primary partition reflects the fact that typologically unmarked front vowels are unrounded, and back vowels are typically rounded. This naturally translates into a separate feature which drives front and back vowels further apart in terms of their vector similarity. Nonetheless, vowel pairs that share the same height such as \textipa{[e]} and \textipa{[o]} retain a fairly high degree of cosine similarity, indicating that our vectors are able to adequately reflect and apply vowel features to the used vowel data set.

\subsection{Dimensionality Reduction}

\begin{figure}[t]
    \centering
    \includegraphics[width=0.5\textwidth]{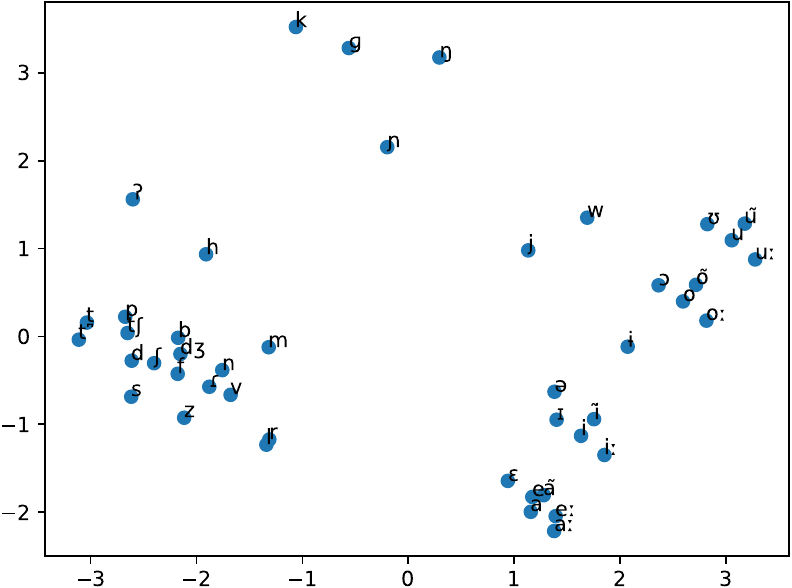}
    \caption{Two-dimensional reduction of feature vectors using PCA.}
    \label{fig:pca}
\end{figure}

\begin{figure}[t]
    \centering
    \includegraphics[width=0.5\textwidth]{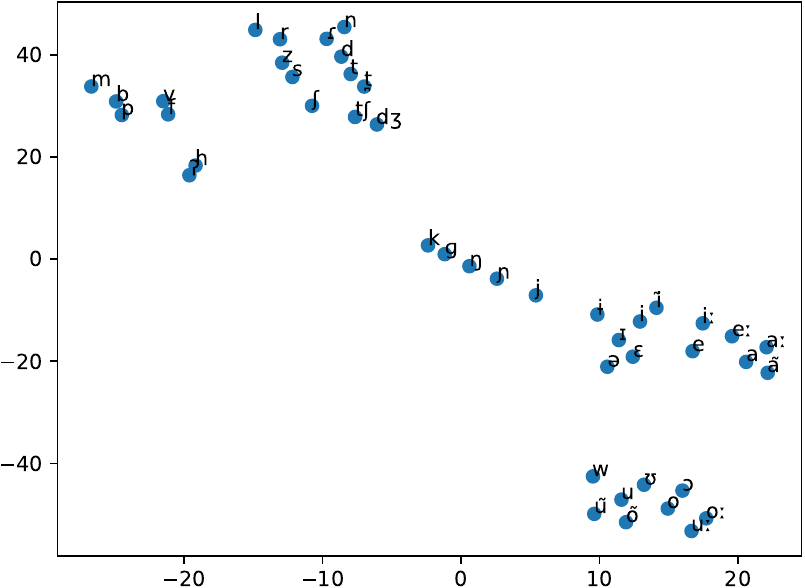}
    \caption{Two-dimensional reduction of feature vectors using t-SNE.}
    \label{fig:tsne}
\end{figure}

We employed two techniques for dimensionality reduction to project the phonological vector space onto a two-dimensional plane, aiming to gain a comprehensive understanding of the inherent structure of our vectors. The employed techniques are \textit{principal component analysis} (PCA; Figure \ref{fig:pca}), known for its ability to reveal global linear structures, and \textit{t-distributed stochastic neighbor embedding} (t-SNE; Figure \ref{fig:tsne}; \citealt{vandermaaten2008visualizing}), known for capturing local nonlinear patterns in the data. Both dimensionality reductions were computed in Python using the SciKit-Learn package \citep{pedegrosa2011sklearn} and visualized with Matplotlib \citep{hunter2007matplotlib}. By employing PCA and t-SNE at the same time, we sought to ensure a robust and detailed exploration of our vectors, leveraging the complementary strengths of both of these dimensionality reduction techniques \citep{Anowar2021dimensionality}.

Figure \ref{fig:pca} and Figure \ref{fig:tsne} visualize the results after dimensionality reduction. Both PCA and t-SNE reveal a consistent narrative, grouping similar sounds together. First and foremost, there is a clear distinction between consonants and vowels, with semi-vowels positioned either between the two clusters in PCA or in close proximity to the corresponding vowel cluster in t-SNE. Once again, this reaffirms the ability of our binary vectors to adequately distinguish between the sounds in a phonologically informed way. 

Focusing on the vowel clusters in Figure \ref{fig:pca} and Figure \ref{fig:tsne}, it becomes evident once again that vowels are primarily divided into (unrounded) front and (rounded) back vowels, aligning with the well-established phonological classification of vowel sounds. Shifting attention to the consonant clusters in both panels, we once more observe that they are primarily grouped by their place of articulation. In both Figures, the velars \textipa{[k,g,N]} form a rather isolated cluster, loosely associated with the palatal nasal \textipa{[\textltailn]}. Both techniques also tend to isolate the glottal sounds \textipa{[h,P]}, with t-SNE placing this sound pair much closer to each other. A distinct picture emerges for the remaining consonants, the coronals and labials: While PCA seems to bundle all of them together into a single large cluster, t-SNE forms two distinct clusters based on their place of articulation, however retaining a certain proximity between the two. The t-SNE plot also exhibits a compelling parallelogram symmetry among quadruplets of stops and fricatives in their voiced and voiceless versions: The alveolars \textipa{[t,d,s,z]} form one such parallelogram in the two-dimensional projection; and a similar pattern can be observed for labials \textipa{[p,b,f,v]}.

The observed similarities depicted in Figure \ref{fig:heatmap-cons} and Figure \ref{fig:heatmap-vow} as well as the patterns in the vowel and consonant clusters depicted in Figure \ref{fig:pca} and Figure \ref{fig:tsne} align with established phonological classifications, providing a visual representation that echoes the theoretical descriptions and classes in phonological theory to a considerable extent. Our observations confirm the potential utility of our phonological feature vectors in computational models, suggesting that the vectors capture meaningful distinctions and relationships inherent in the sounds of human languages.



\subsection{Equivalence Classes}\label{sec:eq-classes}

The current version of CLTS (v.2.1.0) provides a collection of 8,684 unique sounds that were observed in its source datasets. Employing our system, these 8,684 sounds map to 5,285 distinct feature vectors. The system is therefore capable of providing a unique representation for 60.9\% of this large sound inventory, even though it contains a number of very narrow transcriptions, or aspects that we deliberately chose not to represent in the feature space, such as suprasegmental properties being attributed to a segment (for example, putting tones on vowels).

The largest two equivalence classes contain 18 segments respectively, which are all mid and open-mid vowels. The first class therefore contains sounds that are based on \textipa{[@]} and \textipa{[3]}, including among others \textipa{[\'@, @\textrhoticity, \|+3]}. All of these modifications are deliberately disregarded by our system: Tones are suprasegmental features that should not be represented as part of a segment, rhotics lack reliable phonetic correlates \citep{chabot2019rhotic}, and specifying the relative tongue position is an overly narrow transcription style that does not carry distinctive information.

This illustrates the principle of economy, in that we only define as many features as strictly necessary to keep the individual features meaningful and avoid feature inflation. The distinctions that are lost by employing this procedure are extremely narrow in domain, and phonetically not meaningful, as we will argue in the following section.

\subsection{Distinctiveness}\label{sec:distinctiveness}

We investigate the discriminative potential of our system by applying it to all sound inventories observed in phonetic transcriptions of lexical data in Lexibank \citep{list2022lexibank,Lexibank}. The aggregated dataset combines numerous datasets into one unified dataset, spanning over 2,905 language varieties in total. In Table \ref{tab:distinctiveness}, we report the metrics of how well the sound inventories of the languages in Lexibank 1.0 can be described by our system. We report the number of confused sounds per language, that is how many sounds in an inventory share their feature representation with another sound present in this inventory. Formally, this is the difference between the size of the sound inventory, and the number of unique feature representations corresponding to the inventory.

\begin{table}[t]
\centering
\begin{tabular}{@{}ccc@{}}
\toprule
\textit{n} confused sounds & \textit{n} varieties & Portion \\ \midrule
0                         & 2,376               & 0.818      \\
$\leq$1                         & 2,567               & 0.884      \\
$\leq$2                         & 2,648               & 0.912      \\
$\leq$3                         & 2,689               & 0.926      \\
$\leq$4                         & 2,841               & 0.978    \\ \bottomrule 
\end{tabular}
\caption{Number of language varieties in Lexibank 1.0 with at most \textit{n} confused sounds.}
\label{tab:distinctiveness}
\end{table}

The sound inventories of 2,376 varieties, amounting to 81.8\% of all varieties in the dataset, can be represented with full distinctiveness by our system, meaning that every sound is mapped to a unique feature vector. With 2,841 (97.8\%) varieties, the grand share of the dataset's sound inventories can be represented with a maximum of four overlapping feature representations. These overlaps can usually be explained by narrow transcriptions, where the exact realization of the sound is predictable from the context, or can even hint at transcription errors or inconsistencies in the source data.

We investigate such sets of overlapping sounds in their context using concordance lines. This technique is frequently used in corpus linguistics for visualizing in which contexts a certain word appears. Usually, concordance lines are generated by aligning the highlighted target word to the center of a table, and placing the contexts to its left or right respectively \citep[47]{hunston2022corpora}.

\begin{figure}[htb]
    \centering
    \includegraphics[width=0.48\textwidth]{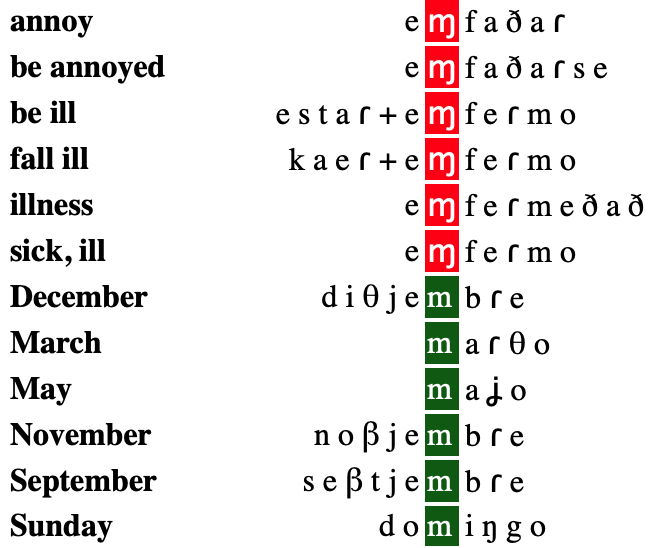}
    \caption{Concordance line for Spanish transcriptions featuring \textipa{[M]} or \textipa{[m]}.}
    \label{fig:cline-spa}
\end{figure}

Figure \ref{fig:cline-spa} exemplifies the usefulness of concordance lines to analyze the contexts in which sounds occur. Here, we investigate the instances of the bilabial and labio-dental nasal consonants \textipa{[m]} and \textipa{[M]} in the Spanish data from the NorthEuraLex database \citep{dellert2020northeuralex}. Both sounds are represented by identical feature vector, but the data makes a distinction between them. Analyzing the relevant forms, however, shows that the presence of \textipa{[M]} can be clearly predicted from the context, since it can only occur before labio-dental obstruents. This suggests that there is no actual distinction between these sounds, since the different surface forms can be explained by a fully predictable assimilation process. Similar cases of complementary distribution can be observed within the same dataset: In Nanai, \textipa{[\|+i]} is only found preceding bilabial and alveolar consonants, while \textipa{[i]} occurs elsewhere; Estonian \textipa{[k]} is transcribed as \textipa{[\r*k]} if preceded by \textipa{[N]}; and Korean voiceless stops are unreleased in word-final position, leading to pairs of distinct transcriptions (e.g. \textipa{[t] - [t\textcorner]}) with a distribution that is completely predictable by context.

\section{Discussion}
In this study, we introduced a new approach to turn the features for all sounds covered by the CLTS reference catalogue into numerical feature vectors. Given that CLTS not only underlies the Lexibank repository, which offers phonetically transcribed, standardized wordlists for more than 2,000 language varieties, but is also used in many additional applicatations that make use of the standards proposed by the CLDF initiative, this means that the binary feature vectors we propose are directly available for a very large number of language varieties.

To assess the effectiveness of our approach, we conducted a detailed analysis using cosine similarity and dimensionality reduction techniques. The resulting similarity patterns, evident in both PCA and t-SNE plots, align with established phonological classifications. Notably, the model distinguishes between vowels and consonants and groups similar sounds based on their place of articulation.
Furthermore, we successfully mapped a substantial inventory of sounds from CLTS to their respective vectors, covering more than half of this extensive sound dataset with unique representations.
Finally, we evaluated the distinctiveness of our vector representations by discerning speech sounds from lexical data in the Lexibank repository. Our system accurately represented a significant portion of the data, ensuring full distinctiveness by uniquely mapping each sound to a feature vector. 

In conclusion, our approach not only provides a practical solution to address general limitations of the pragmatic feature system underlying the CLTS reference catalogue but also offers a flexible approach for representing phonological features in computational linguistics. By converting CLTS feature bundles into binary feature vectors, the approach enables researchers to integrate phonological insights into various computational tasks, ranging from phonology and historical linguistics to linguistic typology.

For the field of cognitive language modeling, our feature system offers an enhanced, more precise phonological representation. \citet{Nieder2024} utilized historical sound class representations in their language processing model \citep[Linear Discriminative Learning, see][]{Nieder2024} to explore mutual intelligibility among Germanic languages. Expanding such models with phonological vector representations instead, may offer new insights into how speech sounds influence meaning and vice versa, thereby guiding language processing and language learning.
 
For historical language comparison, feature representations can be used to dynamically extend fixed-size scoring matrices in computational tasks such as phonetic alignment \citep{kondrak2000new,List2012c} or phonological reconstruction \citep{Bouchard-Cote2013,Jaeger2019,Meloni2021}. While state-of-the-art approaches to phonetic alignment typically deal with the problem of unseen sounds by resorting to sound class representations that represent sounds in phonetic transcriptions in small classes of similar sounds ranging from 10 to about 40 distinct sound classes in total (see \citealt{List2014d} for details on sound class systems), feature vectors would offer a much more fine-grained representation of similarities and differences between individual sounds whose impacts on alignment quality have not been fully tested so far (an exception is the feature-based system by \citealt{Kilani2020}, which requires, however,  sound-feature mappings to be set up manually). For the still unsolved task of unsupervised phonological reconstruction \citep{List2024fca}, a common problem of those approaches that have been proposed so far is that they cannot propose sounds in ancestral languages that have not been attested in the descendant languages. Here, feature vectors might propose a way of handling the unknown, since the vector representation might well propose feature combinations for ancestral sounds that are not observed in individual languages, thus creating unseen sounds from attested sounds. But further tests would be needed to explore the potential of feature vectors in phonological reconstruction.

In summary, we hope that feature vectors, as they have been introduced here, will prove useful in advancing computational approaches in linguistics and integrating linguistic insights into machine learning approaches.


\section*{Supplementary Materials}
Data and code of this study are curated on GitHub (\url{https://github.com/cldf-clts/soundvectors}), the  \texttt{soundvectors} package is also available via the Python package repository PyPi (\url{https://pypi.org/project/soundvectors}, Version 1.0). 

\section*{Limitations}

As discussed in Sections \ref{sec:eq-classes} and \ref{sec:distinctiveness}, we want our system to distinguish sounds that sound differently and avoid lumping them together. Our quantitative and qualitative analyses show that the system seems to be capable of maintaining a high degree of distinctiveness, however, it is not guaranteed that all phonemic contrasts in the world's languages are represented truthfully.
While our phonological feature vectors are a good approximation to spoken language, we want to point out that they cannot perfectly reflect phonetic similarity -- some features are intuitively more meaningful than others, which is not explicitly represented in the vector space; and by extension, similar sounds might differ in a ``disproportionately large number of features'' \citep{kondrak2000new}. \citet{heeringa2004measuring} shows that employing binary features directly as a cost function is not superior to plain edit distance to measure phonetic similarity in dialectal data.

These findings do not undermine the potential of feature systems in analyzing sounds, but rather show that feature vectors should be processed in some way and not be taken at face value. In fact, a number of studies have shown the usefulness of phonological feature vectors for processing sounds in machine learning approaches (e.g. \citealt{staib2020phonological,lux2022language}). This emphasizes the need for a robust system  that reliably generates feature vectors for all IPA segments.

\section*{Acknowledgements}

The authors would like to thank Johannes Dellert for fruitful discussions on the choice of the feature inventory and several feature mappings, as well as two anonymous reviewers for their helpful comments and suggestions. This project was supported by the ERC Consolidator Grant ProduSemy (PI Johann-Mattis List, Grant No. 101044282, see \url{https://doi.org/10.3030/101044282}). Views and opinions expressed are however those of the author(s) only and do not necessarily reflect those of the European Union or the European Research Council Executive Agency (nor any other funding agencies involved). Neither the European Union nor the granting authority can be held responsible for them.




\bibliography{custom}




\end{document}